# Foundation Models in Radiology:
# What, How, When, Why and Why Not

Magdalini Paschali, PhD[1,2], Zhihong Chen, PhD[1,2], Louis Blankemeier, MS[1,3], Maya Varma, BS[1,4], Alaa Youssef, PhD[1,2], Christian Bluethgen, MD, MSc[1,5],

Curtis Langlotz, MD, PhD[1,2,6,7], Sergios Gatidis, MD[1,2], Akshay Chaudhari, PhD[1,2,7]

[1]Stanford Center for Artificial Intelligence in Medicine and Imaging, Palo Alto, CA, USA.
[2]Department of Radiology, Stanford University, Stanford, CA, USA.
[3]Department of Electrical Engineering, Stanford University, Stanford, CA, USA.
[4]Department of Computer Science, Stanford University, Stanford, CA, USA.
[5]Diagnostic and Interventional Radiology, University Hospital Zurich, University of Zurich, Zurich, Switzerland.
[6]Department of Medicine, Stanford, CA, USA.
[7]Department of Biomedical Data Science, Stanford, CA, USA.

**Abstract**

Recent advances in artificial intelligence have witnessed the emergence of large-scale deep learning models capable of interpreting and generating both textual and imaging data. Such models, typically referred to as foundation models, are trained on extensive corpora of unlabeled data and demonstrate high performance across various tasks. Foundation models have recently received extensive attention from academic, industry, and regulatory bodies. Given the potentially transformative impact that foundation models can have on the field of radiology, this review aims to establish a standardized terminology concerning foundation models, with a specific focus on the requirements of training data, model training paradigms, model capabilities, and evaluation strategies. We further outline potential pathways to facilitate the training of radiology-specific foundation models, with a critical emphasis on elucidating both the benefits and challenges associated with such models. Overall, we envision that this review can unify technical advances and clinical needs in the training of foundation models for radiology in a safe and responsible manner, for ultimately benefiting patients, providers, and radiologists.

**Essentials**

- Understanding foundation models in radiology requires clear definitions, insights into their construction, and training methodologies (Pages 2-10).
- Adapting foundation models for radiological applications underscores the need for large-scale, multi-site datasets and highlights the necessity for standardized evaluation benchmarks (Pages 11,14,15).
- Foundation models have various capabilities in radiology, benefiting patients and clinicians and enhancing workflows. Yet, they are associated with risks and challenges that need to be addressed to ensure their responsible deployment (Pages 13-18).

## 1. Introduction

Advancements in artificial intelligence (AI) have led to models that excel in specific tasks, often outperforming humans in controlled environments. For instance, given input radiological images and labels by human experts, traditional AI models have been trained using supervised learning to perform tasks such as disease detection and image segmentation with high accuracy (1, 2). However, such models are often limited by their need for large quantities of labeled data and their inability to adapt to unseen scenarios. To this end, foundation models (FMs) aim to address these challenges. FMs are trained with large-scale unlabeled datasets without the need for extensive expert annotations and can flexibly be adapted across tasks.

FMs are typically large-scale neural network architectures trained primarily on large unlabeled datasets used in natural language processing (NLP) and computer vision. Model architectures, such as the transformer (3, 4), allow for building FMs with billions of trainable parameters by learning from an immense quantity of data. This enables FMs to learn rich data representations, providing a strong starting point for their subsequent adaptations to specific applications. An overview of FMs in radiology is presented in Figure 1.

## 2. What is a Foundation Model?

Amidst the ambiguity surrounding what constitutes a foundation model, it is useful to establish a framework that details key characteristics of FMs. These properties, detailed below, include: 1) incorporating large-scale model architectures and training data, 2) extracting knowledge from multiple data modalities, 3) employing self-supervised training strategies to alleviate the need for extensive expert-labeled datasets, and 4) exhibiting emergent capabilities beyond their training objectives.

Distinct features of FMs, compared to traditional AI algorithms, are their flexibility and efficiency due to their *large-scale architectures and training data*. Model performance follows power laws, consistently improving as model and data size increase (5, 6). In the general domain, FM architectures range from a few billion to over a trillion parameters (7). The largest FMs in the medical domain are adapted from general-use FMs and scale up to 540 billion trainable parameters so far (8). Regarding the scale of FM pre-training datasets, they can consist of over 5B image-text pairs in the general domain with datasets such as LAION-5B (9). Given that larger models require more data to parametrize the model weights, AI models for radiology usually have fewer parameters (6). Within radiology, datasets such as RadImageNet (10) exceed 1M medical images, and MIMIC-CXR (11) consists of over 300,000 pairs of images and radiology reports.

Another feature of FMs is their capability to *process multimodal data* (12). In the general domain, FMs usually process natural images and text. In radiology, multimodality encompasses diverse medical data, including radiological images (X-rays, CTs, MRIs, etc.), reports, clinical notes, electronic medical records, and laboratory findings. In the future, models could expand to omics data and signals from wearable biosensors (12). Combining multiple modalities leverages the interconnectedness of a patient's clinical information, reflects the holistic diagnostic approach of clinicians, and can improve model training by requiring fewer expert annotations (13).

In *self-supervision*, models learn to understand and process data by utilizing objectives from the input data itself rather than relying on labels produced by domain experts (14). An example in radiology is training FMs to match radiological images with their corresponding radiology reports by leveraging the information naturally present in the report to train the image models, and vice versa (8). These data pairs are available in medical databases but typically lack specific labels for training supervised models for diagnostic tasks. This shift from fully-supervised to self-supervised training can alleviate the need for manual annotation and significantly reduce the cost and time to collect radiological labels, which ranges from a few seconds for simpler tasks and up to several minutes per study for fine-grained tasks (15).

Finally, *emergent abilities* in FMs refer to capabilities that manifest as the size of the models grow and are not observable in smaller-scale model versions. Such abilities are not explicitly programmed or anticipated during initial training and show that FMs can generalize to previously unseen concepts and tasks (16). For instance, Med-PaLM-M, an FM trained on multiple biomedical modalities, could predict tuberculosis on Chest X-rays at test time with comparable accuracy to smaller task-specific models without being explicitly trained to identify the disease (8). Similarly, models designed for generating images from text descriptions have shown an ability to perform image classification without task-specific training (17). When thoroughly evaluated and monitored, emergence could contribute positively to the field of radiology, which is constantly evolving with new medical knowledge and technologies (18).

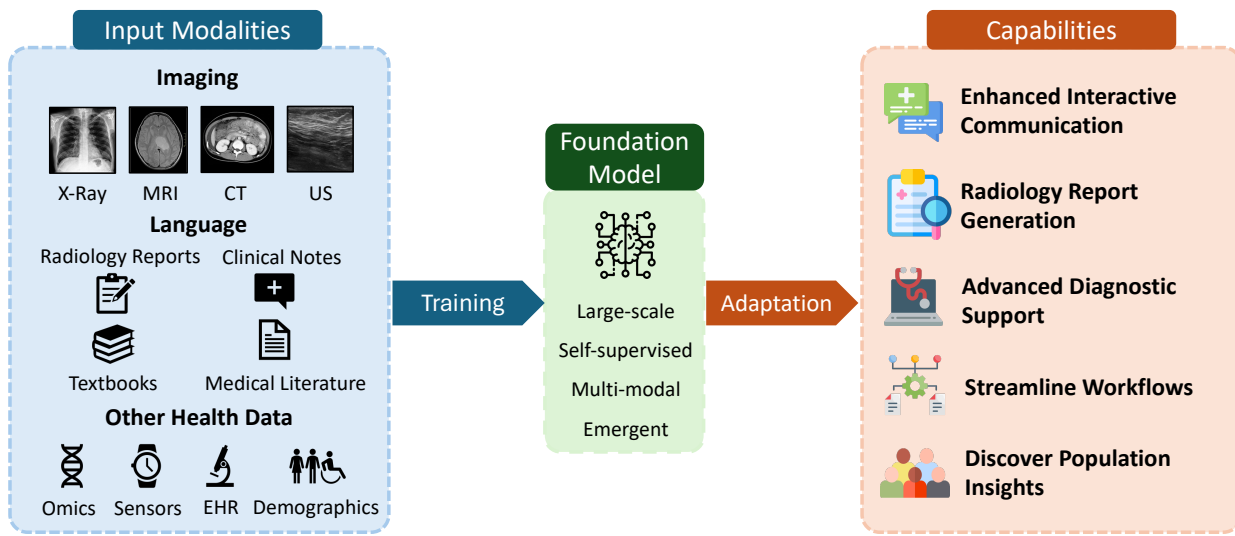

Figure 1: Overview of Foundation Models (FMs) in Radiology along with their inputs, properties, and capabilities. We showcase the various inputs used in radiology FMs, such as multi-modal medical data, ranging from radiological images to electronic health records. Next, we highlight the core FM properties of using large-scale architectures and datasets, knowledge extraction from multi-modal data, and self-supervised learning that minimizes the need for expert annotations. Finally, we list some capabilities of FMs in radiology that extend from enhancing patient communication to unlocking new insights into disease patterns across populations.

### 3. How to Train a Foundation Model?

In this section, we explore the key building blocks and training of FMs.

**1. Modality-specific encoders:** Encoders compress high-dimensional input data into low-dimensional compact representations, called embeddings. These representations facilitate the efficient handling of diverse data types by AI models and compactly encode patterns from the input data. Each modality (e.g., images, text) requires a specialized encoder to transform the raw data into meaningful representations. For example, a vision encoder might convert high-dimensional inputs like images into low-dimensional numerical features describing shape, color, and texture. Similarly, a language encoder turns text into a sequence of vectors representing words and their relationships.

Encoders are trained to extract meaningful features from input data and compress them into low-dimensional embeddings. In a multi-modal setting the objective is to establish a unified understanding across modalities in a shared embedding space. This integration is commonly facilitated through contrastive learning, where the model aligns similar pairs of embeddings, for instance, images from multiple views or images and radiology reports originating from the same patient, while increasing the distance across dissimilar pairs (19). By updating the encoder through this training process, the alignment across modalities is refined, and the encoders learn to extract the most important features from each modality.

**2. Fusion module:** After individual encoders process each modality, fusion modules combine this compact information. This step is performed so that the fused representations can be passed on to the next block, called the decoder, which can perform numerous tasks that benefit from the combined modality information. Though fusion can take many forms, cross-attention allows the fusion of embeddings from different encoders by learning the inter-modality and intra-modality relationships, for example between image regions and words in a sentence (20). Additionally, modality adapters have simplified the process of fusing embeddings. Adapters convert the representations of a modality to a format that can easily be interpreted and used along with the other modalities (21, 22).

**3. Multimodal decoders**: Decoders transform the low-dimensional compact representations created by encoders and fused across modalities back into high-dimensional outputs suitable for various tasks with different output dimensionalities. Depending on the desired output and task, the model might need specialized decoders for each modality. For example, an image captioning task might use a language decoder to generate text (23). In contrast, a visual question-answering task might employ a vision decoder to highlight relevant parts of an image based on a question (24).

Next, we present an overview of encoder training styles, specifically generative and contrastive pre-training. Generative pre-training teaches models to generate new data, enhancing the models' understanding of data distributions. Contrastive pre-training sharpens the ability of FMs to discriminate between similar and dissimilar samples, improving pattern recognition. For each subsection, we first describe pre-training for language models, followed by vision models.

### 3.1 Generative Pre-training

Generative learning allows models to understand the underlying distribution of input data and improves an FM's generalization by learning to reconstruct crucial data features.

**Auto-regressive models** are adept at predicting data to match a specific distribution and capturing dependencies in sequential data. By predicting the next element in a sequence based on previous elements, they model the sequential relationships within the data. In NLP, models like Generative Pre-trained Transformer (GPT) (25) have advanced this approach, generating text with unprecedented coherence and context understanding. For computer vision, autoregressive modeling can be extended to sequentially predict the pixels in an image along two spatial dimensions (26), as shown in Figure 2.

**Autoencoders**, another generative approach, are designed to compress inputs into a compact representation and reconstruct the original input from this representation. This process helps models learn the inherent structure of the input data. Variational Auto-Encoders (VAEs) (27), a specialized form of autoencoders, introduce a probabilistic approach by generating a distribution over the representation space. This allows for generating new data (e.g., images) and facilitates anomaly detection by evaluating how well new inputs fit within the learned distribution. By learning distributions rather than fixed representations, VAEs can capture a broader understanding of the data's underlying structure. Diffusion Models (DMs) (28, 29) represent another approach to generative learning. DMs transform the input data by gradually adding noise over several steps, then learn to reverse this process, effectively performing 'denoising' to recreate the original data. This iterative approach allows for generating detailed outputs, making DMs particularly adept at creating realistic images and simulating complex data distributions. Finally, text-conditioned DMs for medical imaging can create clinically relevant images guided by textual input (30, 31).

Another concept of generative pre-training is learning from incomplete information, which involves deleting ("**masking**") parts of the input data and training the model to generate the masked regions. In NLP, masking is used in Bidirectional Encoder Representations from Transformers **(**BERT) (32), where parts of the text inputs are masked and the model is trained to predict the hidden words. This strategy enables models to grasp the context and semantics of language, gaining a deeper understanding of textual structure. Similarly, in computer vision, Masked Autoencoders (MAE) (33) mask patches of images, training the model to reconstruct the missing parts. This approach enhances model robustness to partial information and improves feature extraction. In medical imaging, Medical MAE (34) is trained to reconstruct partially masked images, increasing classification and segmentation accuracy in 2D and 3D scans (35).

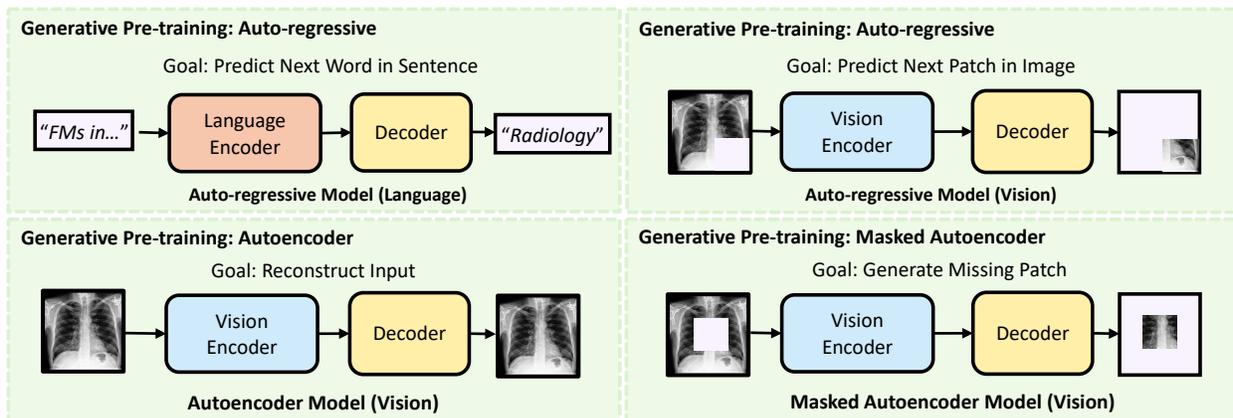

Figure 2: Generative Pre-Training Techniques. Models are trained to reconstruct or complete data, enhancing the models' understanding of spatial context. (1) Top Left: Auto-regressive models for natural language processing, predicting the next word in a sentence to extract knowledge of linguistic patterns. (2) Top Right: Autoregressive models for computer vision, predicting the next patch within chest X-rays to enhance the spatial data comprehension of FMs. (3) Bottom Left: Autoencoders, which learn to compress and reconstruct inputs, highlighting their ability to learn the intrinsic structure of images. (4) Bottom Right: Masked Autoencoders, focusing on reconstructing occluded sections of images, training FMs to be robust to incomplete information.

### 3.2 Contrastive Pre-training

Contrastive learning trains a model to discriminate between similar and dissimilar input samples. The goal is to minimize the distance between similar input samples (attract positive pairs) and maximize the distance across dissimilar representations (repel negative pairs) as shown in Figure 3. In medical imaging, X-rays from different views or corresponding scans and radiology reports originating from the same patient can constitute a positive pair, while scans and reports from different patients are negative pairs. Contrastive learning can be used on representations from one modality, like imaging, but also across modalities to help models achieve alignment across, for instance, images and text.

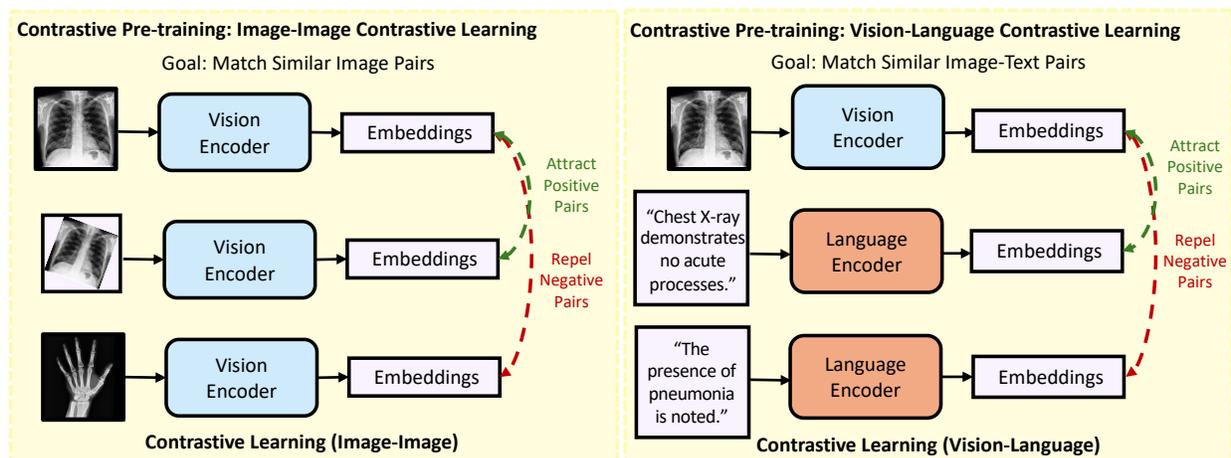

Figure 3: Contrastive Pre-Training. Models learn to align similar pairs of embeddings while increasing the distance across dissimilar pairs. (1) Left: Image-image contrastive learning, showcasing a chest X-ray and its rotated version as a positive pair of similar samples and a hand X-ray to illustrate the negative pair. (2) Right: Vision-language contrastive learning, contrasting a chest X-ray with one matching and one non-matching radiology finding. This method trains models to understand images in the context of descriptive text by matching relevant images and textual information.

One approach to generating pairs of images for contrastive learning is applying different transformations (cropping, flipping, etc.) to an input, creating two views. These views of the same image are a positive pair, and views of different images constitute a negative pair. Methods such as the Simple Framework for Contrastive Learning (SimCLR) (36) train models to minimize the distance across positive pairs and maximize the distance across negative pairs. Other approaches, such as Bootstrap your own latent (BYOL) (37), achieve the same goal but rely solely on positive pairs by comparing two transformed views of the same image, and bringing the representations of the two views closer in the representation space (38).

**Vision-language contrastive learning** has advanced the pre-training of FMs since language supervision can benefit computer vision tasks by leveraging the semantic relationships between images and text (39-41). In Contrastive Language–Image Pretraining (CLIP) (42), models learn to minimize the distance between corresponding image-text pairs while maximizing the distance across non-matching image-text combinations. CLIP enables models to understand images in the context of descriptive text. This capability is especially beneficial for radiology, where interpreting images alongside reports or annotations is routine. MedCLIP (43) extends the positive image-text pairs beyond a single patient and constructs positive pairs across patients using scans and text with high semantic similarity based on image disease labels and text extracts (44).

### 3.3 Adapting FMs for Task-Specific Applications

Having discussed the pre-training approaches that create robust encoders, we will now explore how these pre-trained building blocks can be adapted to meet the demands of clinical applications, as summarized in Figure 4.

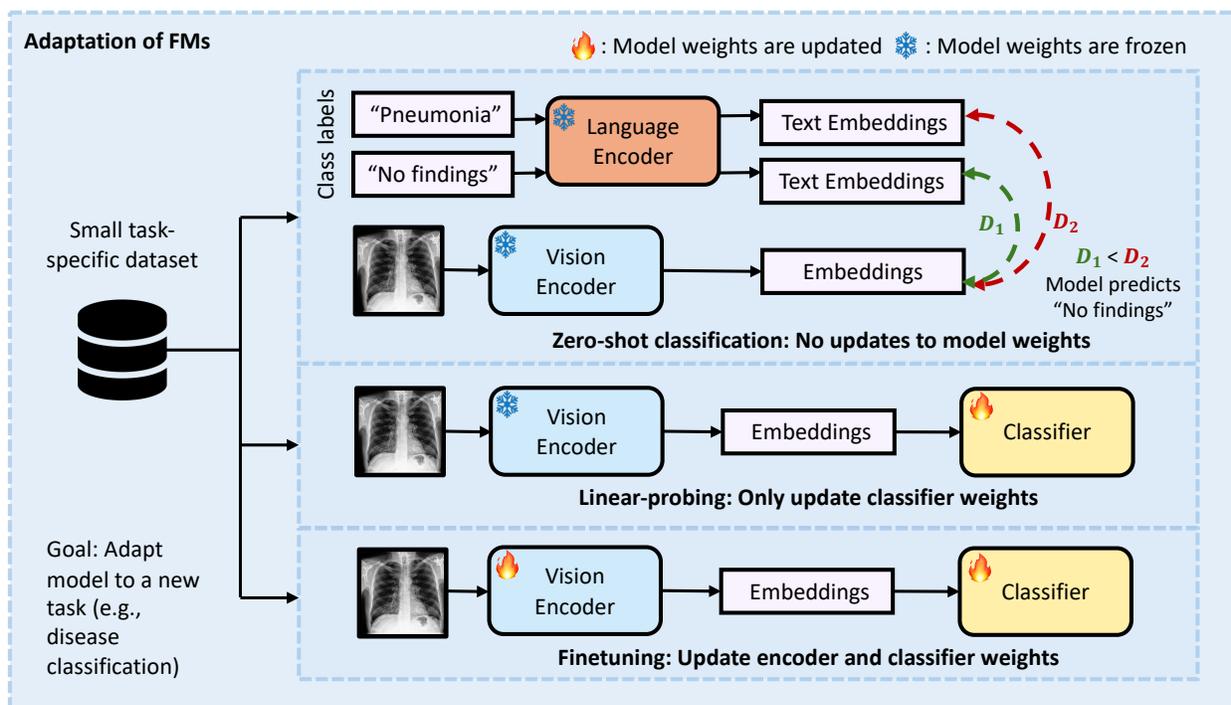

Figure 4: Adapting Foundation Models to Clinical Tasks. After the encoder pre-training, FMs can be adapted to specific clinical applications using several methods. (1) Top: Zero-shot classification is illustrated by comparing a chest X-ray with two classes (pneumonia and no findings). We use textual class descriptions to match the image embedding to the closest text embedding without updating the model. The predicted class is determined based on the text embedding with the shortest distance to the image embedding. (2) Middle: With linear probing the weights of the model encoder are not updated, and an extra classifier layer is trained using a small dataset for a novel task. This approach balances pre-trained knowledge and adapting FMs to specific requirements.

(3) Bottom: Fine-tuning updates both the model's encoder and the added classifier using a small, task-specific dataset, fully adapting to the nuances of a new clinical task.

**Zero-shot or few-shot adaptation** requires zero to few examples to adapt a pre-trained FM to new tasks. In zero-shot inference, a model is applied to a new task without using any annotated examples; in few-shot learning, the model is only given a few examples to learn from (45, 46). Suppose a model was trained on X-rays with attributes like "calcification," "mass," or "increased bone density." During training these features are associated with known diseases, such as bone tumors or osteoporosis ("reduced bone density"). In zero-shot learning, the model can identify a rare bone disease not seen during training by recognizing familiar attributes. In this setting, CLIP-style pre-trained models can classify images without task-specific training by transforming the classification task into an image-text matching problem. Class names are converted into descriptive sentences (e.g., the class *pleural effusion* becomes "an image of pleural effusion"), allowing FMs to leverage their attribute understanding acquired during pre-training.

**Task-specific linear probing/fine-tuning** involves adapting a pre-trained model to a specific radiology task, such as diagnosing from X-rays or MRIs, by training it further on a smaller, task-specific dataset. This process fine-tunes the model's parameters to align closely with the diagnostic patterns of medical imaging. In cases where the encoders might be too large or proprietary, linear probing can be used to adjust only an additional final layer of a model to new tasks, preserving the learned representations in the earlier stages of the model (47).

**Instruction tuning** is a form of fine-tuning where a pre-trained model is further trained on a dataset that consists of instructions paired with their corresponding outputs (48, 49). An example is "given an *image* - generate the *radiology report*". The goal is to enable the model to understand and execute various tasks based on natural language instructions. During this process, the model weights, including those in the encoders, are adjusted to minimize the difference between the model's outputs and the expected outcomes as defined by the instruction-task pairs (50-52). This is particularly useful in radiology, where models can be instructed to identify or quantify pathological features from images (53), (22). After a model has been instruction-tuned, **chain-of-thought prompting** can further guide it in reasoning language tasks (54). This method refines the model's utility to articulate intermediate steps while formulating a conclusion. This could enhance interpretability by outlining the rationale behind a particular decision (55).

Lastly, **Reinforcement Learning from Human Feedback (RLHF)** (56) involves training a model using reinforcement learning, where the model receives rewards for its actions that are based on human preferences. This process allows the model to adjust its outputs to better align with human judgments (57, 58).

## 4. When to Expect Radiology FMs?

In developing FMs for radiology, leveraging large-scale, diverse datasets is the foremost critical need for pre-training robust models for analyzing complex tasks. In this section, we will delve into the largest publicly available datasets used for FM pre-training and tuning and discuss the requirements for future datasets.

### 4.1 Radiological datasets suitable for FM pre-training

Vision: Current radiological datasets include up to 1M automatically or manually annotated CT, MRI, and US scans (10) and a few hundred thousand chest X-rays, that can be used for disease identification (59), (60).

Vision-Language: MIMIC-CXR features over 300,000 chest X-rays paired with free-text radiological reports (11). Additional collections include more than 1M image-text pairs extracted from figures and captions from PubMed medical papers containing various radiological modalities (61). Even though such datasets enable vision-language pre-training, the quality of the provided text scraped from literature is lower than radiological reports. Training exclusively on such data should be performed cautiously since lower-quality pre-training data can provide incomplete or incorrect textual information.

Segmentation: Existing datasets provide more than 4M of medical images and corresponding masks, spanning various anatomical regions (62), supporting the development of segmentation capabilities within FMs. Recently released datasets (63) consist of around 20,000 CXRs expertly annotated for pneumothorax, acute rib fracture, and chest tubes accompanied by free-text reports.

Future datasets must address several needs to foster the integration of FMs in radiology. Firstly, there's a growing demand for 3D/4D datasets, comprising MRI and CT scans, alongside an expansion in datasets for US. Additionally, incorporating radiology reports and imaging data is crucial to represent the context and findings accurately. Moreover, curating longitudinal datasets will allow FMs to capture disease progression over time. Expanding the range of available anatomical regions is also critical, moving beyond thoracic-focused datasets. Given variations in imaging acquisition and quality across sites, there is a need to release datasets with multi-site data. Furthermore, releasing patient demographics, such as age, sex, and race, will enable the development of fairer training approaches and allow for thorough model evaluation. Lastly, ensuring that datasets adequately represent diverse populations is necessary to pave the way for equitable FMs in radiology.

### 4.2 Radiological datasets suitable for FM instruction tuning

Instruction tuning datasets are created for finetuning FMs in radiological tasks by understanding and responding to complex questions (53), (64), (22). For multimodal FMs, these datasets usually consist of data triplets—comprising an image input, a related question, and the corresponding answer—that train the model to adjust to clinical tasks (49). Rather than being built from scratch, they typically combine smaller task-specific datasets and repurpose datasets used for pre-training, such as MIMIC-CXR (11), transformed into an instructional format.

### 5. What Tasks Can Radiology FMs Support?

In this section, we describe how FMs offer a range of capabilities that can benefit both patients and clinicians due to their versatility and multimodal understanding.

**Enhanced Patient Communication**

FMs can be applied in radiology to transform medical reports into patient-friendly language (65). This adaptation increases accessibility, enabling patients to better understand their diagnoses and treatment recommendations (66). Additionally, FMs can translate reports into various languages, catering to diverse patient populations. By easing communication, FMs can bridge the gap between the patient's home and the hospital, alleviating patient anxiety and enhancing comfort.

**Radiological Report Generation**

FMs can generate reports directly from imaging data (22), perform report coding into different diagnostic codes, and summarize key findings concisely and accurately (67, 68). This ensures report consistency and allows radiologists to dedicate more time to complex cases. Moreover, FMs can standardize report quality by identifying inconsistencies or omissions and enhancing clarity and completeness in reports (69). Finally, interactive dialogue facilitated by FMs allows radiologists or patients to ask specific questions based on visual data, improving understanding and communication.

**Advanced Diagnostic Support**

By analyzing multimodal radiological data, FMs can propose diagnoses and suggest treatment plans, aiding in faster decision-making (8, 70). Furthermore, FMs like Medical Segment Anything Model (71) excel at segmenting structures and quantifying volumes, crucial for treatment planning and monitoring disease progression. Another task FMs can assist with is patient triaging. Automating case prioritization based on severity and urgency improves response times for critical cases, ensuring timely intervention (72). This can be particularly beneficial towards improving access to healthcare for regions with limited access to clinical experts (73).

Another application of FMs in radiology includes automating follow-up care. Utilizing imaging reports, FMs could generate personalized messages with recommendations for further care, thereby improving the likelihood of patients completing necessary follow-up procedures. This automation extends to tasks such as information routing within radiology subspecialties and generating templated notes for referring providers. By handling these routine tasks, FMs can allow radiologists to concentrate on direct patient care.

**Streamlined Workflow and Data Management**

FMs can efficiently manage vast databases of images and reports, simplifying information archiving and retrieval. Moreover, real-time image quality monitoring by FMs can alert technologists to imaging artifacts or suggest protocol adjustments for optimal imaging based on diverse patients.

**Unlocking Population Insights and Disease Prediction**

By analyzing vast amounts of data, FMs can identify patterns and biomarkers relevant to public health, aiding in early detection and prevention strategies (74). Furthermore, analyzing longitudinal data enables FMs to predict disease progression, assisting in scheduling timely follow-ups and interventions.

### 5.1 Methods for Evaluating Clinical FMs

After discussing the capabilities of radiological FMs, it is critical to explore strategies to evaluate clinical FMs and understand how radiologists play a pivotal role in both assessing and enhancing FMs in radiology (22), (75), (53).

Given the versatility of FMs in handling multiple tasks and modalities, evaluation frameworks must assess a wide array of capabilities. Specifically, for discriminative tasks (e.g. classification, segmentation), it is crucial to gauge the performance of FMs in disease prediction. This involves assessing the models' ability in coarse-grained settings, such as separating individuals with and without specific findings, and in fine-grained tasks, such as classifying multiple diseases with increasing complexity. Evaluating multimodal FMs is usually performed given an input radiological scan and a prompt that can have the form of a multiple choice question such as "Does this X-ray contain cardiomegaly?" and possible answers "Yes/No" in a binary classification scenario (22). Additionally, tasks that evaluate visual question-answering and image reasoning are important, as they demonstrate a model's ability to detect a disease, associate it with an image region, and provide a rationale for its conclusions.

Assessing the similarity between generated content and ground truth for generative tasks, including text or images, is crucial. However, this is challenging, especially when ground truth is unavailable and since the generated sample can be correct despite not being identical to a given ground truth. Domain-specific benchmarks can evaluate the visual and textual understanding and generation abilities of FMs for various anatomies and modalities (78). Moreover, human evaluation plays a pivotal role in rating generated samples' completeness, conciseness, and correctness, such as radiological reports (76). Moreover, large language models like GPT-4 (25) can be used to compare the accuracy of FMs' generated outputs.

Moreover, a realistic evaluation technique allows radiologists to interact with an FM using a demo to identify failure cases and model weaknesses. Radiological FMs face challenges in generating precise quantitative measures for distances or size of pathological growths and lack understanding of temporal changes due to limited longitudinal training and datasets (22). Overcoming these obstacles is imperative for advancing FMs that are technically sophisticated and clinically valuable. Finally, evaluating bias is also a component of a comprehensive evaluation, helping uncover limitations that might impact a model's effectiveness for specific patient groups (77).

## 6. Cautions for Using FMs in Radiology

Despite the promising capabilities of FMs several challenges exist that need to be addressed to realize the potential of these models in radiology. A summary is presented in Figure 5.

1. Technical and Development Considerations

FMs are not immune to generating errors or "mistakes of fact" in their outputs. Responses from these AI models may be unreliable or inconsistent (79). To mitigate these risks, training FMs on large, and representative datasets with continuous oversight from medical professionals is important. Moreover, AI-generated text may produce hallucinations, leading to inaccurate or incomplete medical reports (80). This signifies the need for standardized evaluation benchmarks assessing the reliability and accuracy of FMs (76).

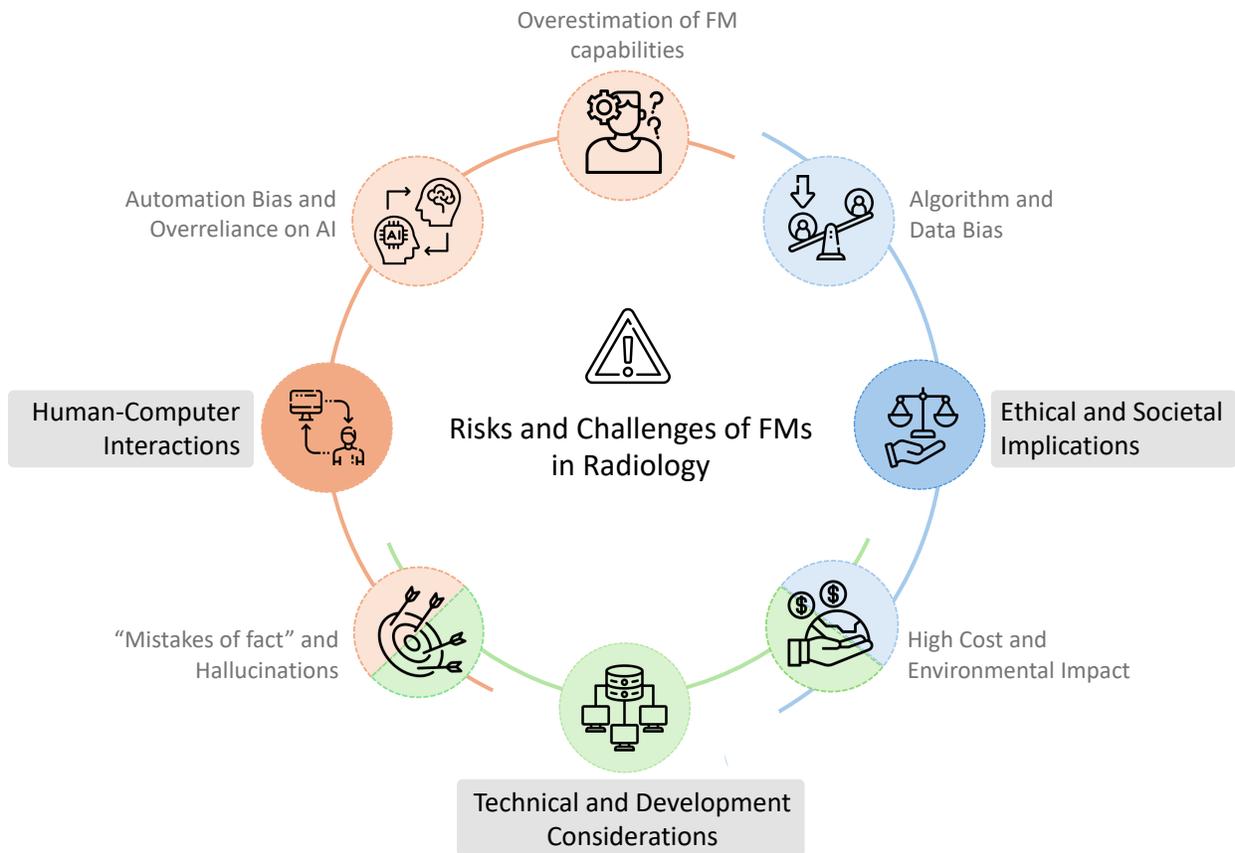

Figure 5: Challenges in Implementing Foundation Models in Radiology. We illustrate the cautions for using FMs in radiology, categorized into three main categories. Technical and Development Considerations highlight the importance of comprehensive training that prevents inaccuracies and "hallucinations" in AI-generated content. Moreover, efficient training and deployment strategies should be developed to lower the financial and environmental impact of FMs. Human-computer interactions stress the need for caution against overreliance on AI, advocating for robust safeguards and clear communication to manage expectations around AI capabilities. Ethical and Societal Implications address the risks of societal bias and data disparity, underscoring the critical need for equitable algorithm design and deployment strategies.

2. Human-Computer Reaction with FMs

As FMs become increasingly trusted in real-world applications, there is a risk of overreliance on AI (81). This can result in automation bias, potentially leading to incorrect diagnoses or treatment recommendations. A similar concern is the tendency to anthropomorphize AI, leading to overestimating its capabilities (82, 83). In a medical setting, this risk is magnified when the readers of the AI output are patients who cannot quickly identify false information.

Therefore, it's essential to implement safeguards and provide clear warnings, emphasizing that FM outputs do not constitute medical advice.

3. Ethical and Societal Implications of FMs

Similar to traditional AI models, FMs can manifest societal bias, evident in data disparities affecting specific population groups (84). Lack of access to healthcare, restrictive clinical trial criteria, and systemic discrimination can lead to skewed data, impacting model fairness (77). Moreover, algorithm design can affect how sensitive attributes are considered and how outcomes are defined and measured. These choices can exacerbate disparities, but they can also alleviate biases when designed appropriately (85). Therefore, evaluating FMs on large, representative populations and performing subgroup analysis is necessary before clinical deployment (86, 87).

Finally, due to their complexity and scale, the hardware requirements to train and deploy FMs pose a financial and environmental challenge (88). This requirement impacts the feasibility of their use in hospital settings and contributes to centralization in a few well-resourced organizations, limiting the diversity of perspectives in AI development.

## 7. Future Directions

Future research in FMs involves optimizing these models to operate on less resource-intensive hardware during training and deployment. This advancement would broaden the accessibility of FMs, allowing smaller healthcare facilities to benefit from these technologies. Efficient FMs would require refining model architectures to maintain high performance with reduced model size, through methods like pruning or quantization (89, 90).

Another area of focus is understanding the dataset size necessary to achieve state-of-the-art results. The performance of FMs varies based on the dataset's size and quality. Future research should explore strategies for efficient data collection, and usage (91, 92). This includes developing techniques to enhance the performance of models trained on limited data, ensuring they can still deliver high accuracy in clinical settings (93).

To this end, advanced data augmentation and synthetic data for training and evaluating FMs are potential areas for future exploration (94). These methods could address the challenges of limited datasets, enabling the creation of diverse training materials without compromising patient privacy or relying on extensive real-world data collection. Future developments should also prioritize privacy-preserving techniques such as differential privacy and federated learning (95) to help maintain the confidentiality of patient data while still allowing for the development and sharing of robust FMs.

Moreover, incorporating continual learning into FMs to adapt in real-time to new data and evolving clinical practices is critical for future applications in radiology (96). Continual learning is the ability of a machine learning model to acquire new knowledge while retaining previously learned information (97). This adaptability is essential in the dynamic healthcare field, where new treatments, technologies, and disease variants continually emerge. Furthermore, continual monitoring and evaluation of deployed FMs are essential for ensuring their ongoing accuracy and fairness in changing real-world conditions (98). This process involves tracking performance metrics, detecting data drift, and updating models to adapt to new data patterns. Feedback mechanisms help identify performance issues and areas for improvement, monitor bias, and ensure regulatory compliance. A combination of automated systems and human oversight is crucial in maintaining the integrity of AI systems over time.

The future development of FMs in radiology requires integrating interdisciplinary knowledge from medical experts, ethicists, data scientists, engineers, and researchers. This collaborative approach ensures that models are technically proficient, ethically sound, and aligned with clinical needs.

## 8. Conclusion

The future of FMs in radiology will be marked by advancements that enhance accessibility and efficiency. By focusing on optimizing models for diverse hardware environments, understanding dataset requirements, preserving patient privacy, and integrating interdisciplinary knowledge, the field can advance toward creating more inclusive and effective AI tools. These developments will pave the way for FMs to play an increasingly integral role in transforming healthcare delivery and improving patient outcomes.

## 9. Summary Statement

We elucidate the fundamentals, construction, and training of foundation models in radiology, underscore the critical need for large-scale datasets and evaluation benchmarks for their adaptation, and explore their potential benefits and inherent risks, guiding towards their responsible deployment in radiological practices.